\documentclass[11pt,letterpaper]{article}
\usepackage[T1]{fontenc}
\usepackage[utf8]{inputenc}
\usepackage{lmodern}
\usepackage[margin=1in]{geometry}
\usepackage{amsmath,amssymb,amsthm,mathtools}
\usepackage{graphicx,booktabs,array,microtype}
\usepackage{caption}
\usepackage[section]{placeins}
\usepackage{flafter,needspace,wrapfig,float}
\usepackage[hidelinks]{hyperref}
\usepackage{xurl}
\usepackage{fancyhdr}
\hypersetup{pdftitle={When Should AI Overturn a Human Decision?
Controversy at the 2026 World Cup and a sqrt(2/pi) Solution},
 pdfauthor={Amir Ali Ahmadi; Parsa Alavi; Radeen Alavi},
 pdfsubject={Gaussian human--AI decision fusion}}
\numberwithin{equation}{section}
\newtheorem{theorem}{Theorem}[section]
\newtheorem{proposition}[theorem]{Proposition}
\newtheorem{lemma}[theorem]{Lemma}
\theoremstyle{definition}

\theoremstyle{remark}

\newcommand{\R}{\mathbb{R}}
\newcommand{\E}{\mathbb{E}}
\newcommand{\Prob}{\mathbb{P}}
\newcommand{\HH}{\mathrm{H}}
\newcommand{\AI}{\mathrm{AI}}
\newcommand{\ML}{\mathrm{ML}}
\newcommand{\sdH}{\sigma_{\HH}}
\newcommand{\sdA}{\sigma_{\AI}}
\newcommand{\mH}{m_{\HH}}
\newcommand{\mA}{m_{\AI}}
\DeclareMathOperator{\Var}{Var}
\DeclareMathOperator*{\argmax}{arg\,max}
\graphicspath{{figures/}}

\title{When Should AI Overturn a Human Decision?\\
\large Controversy at the 2026 World Cup and a $\sqrt{2/\pi}$ Solution}
\author{
Amir Ali Ahmadi\\
\small Princeton University\\
\small\texttt{aaa@princeton.edu}
\and
Parsa Alavi\\
\small Algonquin Regional High School\\
\small\texttt{parsialavi@gmail.com}
\and
Radeen Alavi\\
\small Algonquin Regional High School\\
\small\texttt{radeenalavi@gmail.com}}
\date{}

\makeatletter
\renewcommand{\@maketitle}{%
  \newpage\null\vskip-18pt
  \begin{center}%
    {\LARGE\@title\par}%
    \vskip0.8em
    {\large\lineskip0.5em
      \begin{tabular}[t]{c}\@author\end{tabular}\par}%
  \end{center}%
  \vskip0.5em
}
\makeatother

\begin{document}
\maketitle
\begin{abstract}
We study how to combine a human binary call (a referee’s offside or onside decision) with an AI numerical measurement (VAR’s estimate of an attacker’s position relative to the offside line). We derive an explicit maximum-likelihood rule that determines when the referee’s call should stand and when a conflicting VAR measurement should overturn it. Our model treats the true offside margin as fixed and unknown and assumes independent additive Gaussian errors in the human’s perception and the AI measurement. When both errors have zero means and known standard deviations \(\sigma_{\mathrm H}\) and \(\sigma_{\mathrm{AI}}\), respectively, the reversal threshold is \(\sqrt{2/\pi}\,\sigma_{\mathrm{AI}}^2/\sigma_{\mathrm H}\). An onside call is overturned precisely when VAR’s measured offside margin exceeds this threshold—not merely because VAR places the attacker beyond the offside line. We also derive the rule when either source has a systematic bias. We prove that the maximum-likelihood estimate of the true margin exists and is unique, while the final binary decision requires only a threshold comparison, without computing that estimate numerically. We illustrate the rule through numerical examples and show how to estimate the error parameters from calibration data. The same analysis applies beyond soccer to binary classification problems in which a human call and an AI measurement satisfy the stated model. This paper is intended for educational purposes: we aim to illustrate how undergraduate students and advanced high school students familiar with basic probability and calculus can investigate research questions in AI and its applications.
\end{abstract}

\section{Introduction}\label{sec:intro}

\begin{wrapfigure}{r}{.36\textwidth}
\vspace{-8pt}
\centering
\includegraphics[width=.64\linewidth]{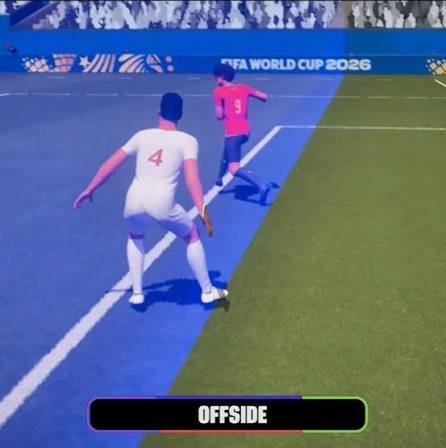}
\caption{The 3D reconstruction of Khalilzadeh’s late winner. According to VAR, the tip of his foot was in an offside position, disallowing the goal.}
\label{fig:iran-intro}
\vspace{-6pt}
\end{wrapfigure}

The 2026 FIFA World Cup, unprecedented in scale, was marked by both extraordinary excitement and considerable controversy. Entire nations were lifted
to their feet with last-minute goals, only for the video assistant referee (VAR) to be the bearer of disappointment
at an agonizingly inconvenient time.

On June 26, 2026, in the final round of group-stage matches, Iran faced Egypt in what was the
most important match either of these sides had played in years. It was a simple
scenario; Iran needed a win to guarantee their qualification to the Round of 32, while
a tie would leave their fate up to other results the next day. Egypt had already
qualified to the Round of 32, but a loss against Iran would put them up against a
European giant as opposed to a weaker side such as Australia. In a packed Seattle
Stadium, with the authors among those in attendance, both teams traded early goals in the first half. The game remained deadlocked into the final twenty minutes, when momentum began to shift. Iran launched a
sustained attack that forced the Pharaohs to sit in a low block for the remainder of
the game. In the 93rd minute, Iran seemed to have finally found the winner. A
free-kick from star full-back Ramin Rezaeian led to a scuffle in the box, resulting in
a Shoja Khalilzadeh goal that lifted the spirits of an entire nation. Celebrations
erupted, as Iran appeared to have made history. Yet the joy soon gave way to disappointment. After minutes of deliberation amid Egyptian
protest, the referee turned to VAR and disallowed the goal for an alleged offside\footnote{Law~11 of the International Football Association defines an offside position by the conditions that any part of the head, body, or feet is in the opponents' half (excluding the
halfway line) and is nearer to the opponents'
goal line than both the ball and the second-last opponent. The hands and arms
of players are not considered~\cite{ifab-offside}.} offense~\cite{reuters-match}.
Although the on-field refereeing team initially ruled the goal onside, the decision was overturned after VAR claimed that a small part of Khalilzadeh’s foot was in an offside position at the relevant moment. Figure~\ref{fig:iran-intro} shows the reconstruction displayed during the review.


This VAR decision proved costly for Iran. Their fate was now up to matches the next day, their Round of 32 dreams dependent on at least one of the following outcomes occurring: Ghana beating Croatia, Uzbekistan not losing to Congo, and the bout between Austria and Algeria having a winner. According to The
Athletic, ``Team Melli'' had a 92\% chance of qualifying for the knockouts despite their
seemingly unjust draw with Egypt~\cite{athletic-tracker}. This probability dropped all
the way down to 0\% as none of the three subsequent results went Iran’s way.
One can argue that Iran’s woes could be traced back to that decisive goal disallowance by VAR. This raises a natural question: should a very small offside margin reported by VAR be sufficient to overturn the on-field referee’s call?


\begin{wrapfigure}{r}{.48\textwidth}
\vspace{-8pt}
\centering
\includegraphics[width=\linewidth]{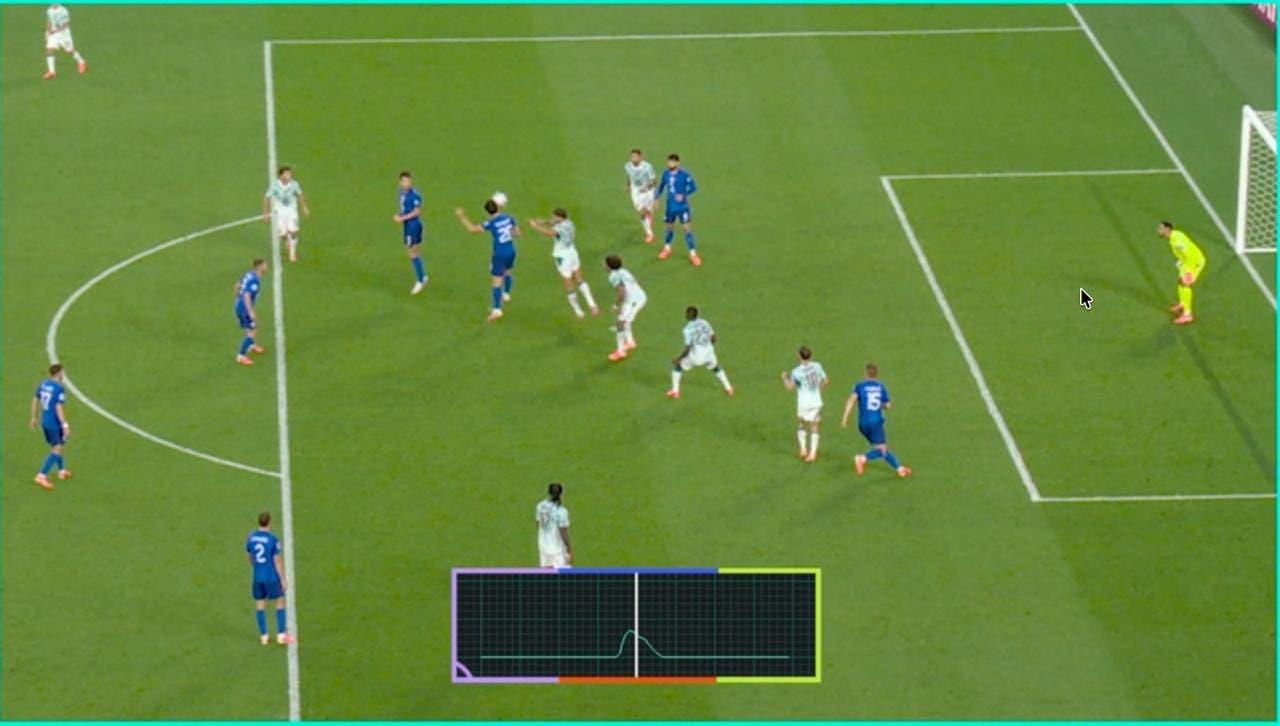}
\caption{The signal used by VAR to support its claim that the ball touched Matanović’s hair. The Croatian federation filed a complaint with FIFA challenging VAR's decision to overturn the human ruling.
}
\label{fig:portugal-intro}
\vspace{-6pt}
\end{wrapfigure}


The Iran–Egypt incident was by no means the only VAR controversy at the 2026 World Cup. During the Portugal-Croatia game, there were multiple disallowed goals, the most
notable of which came in the last minutes of stoppage time. In the 90+13th minute, Joško Gvardiol scored to equalize for his country and force the match into extra time.
However, after lengthy deliberation, Croatian player Mario Pašalić was deemed offside during
the build-up and thus the goal was disallowed. The sequence consisted of an Ivan
Perišić cross into the box followed by a deflection off Portuguese player Renato
Veiga. The ball fell to Pašalić, who failed to control it, but it bounced off him to
Gvardiol, who tapped it in. However, before Veiga’s deflection, there was a purported
touch from Igor Matanović. At the time of this perceived touch, Pašalić was in an offside position and subsequently became involved in the play that led to the goal. Although the alleged contact was nearly imperceptible to viewers watching on television, the sensor within the Trionda ball registered a signal that VAR interpreted as indicating contact with Matanović’s hair~\cite{goal-croatia}. 
Figure~\ref{fig:portugal-intro} shows the sensor trace displayed for this incident. More generally, such incidents illustrate that even highly precise spatial reconstructions need not imply equally precise offside margins: in video recorded at 30 frames per second for example, an athlete moving at \(5\) m/s travels about \(17\) cm between consecutive frames, so even small uncertainty in the relevant contact time can be consequential.



\begin{figure}[H]
\centering
\includegraphics[height=1.80in]{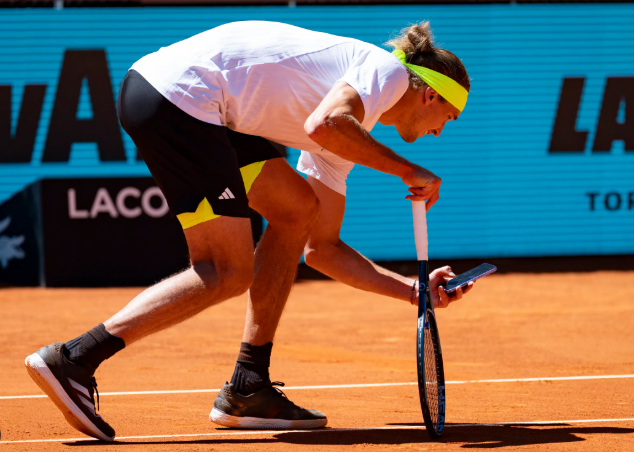}\hspace{.02\linewidth}%
\includegraphics[height=1.80in]{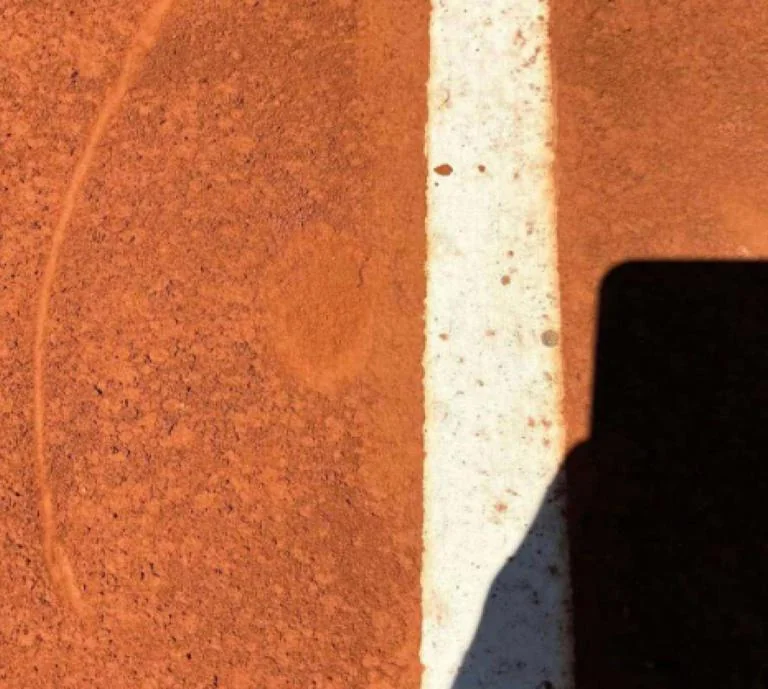}
\caption{Left: Alexander Zverev photographs a ball mark with his phone during the match to challenge an electronic line call that ruled Davidovich Fokina’s shot in. Right: the photograph, which Zverev later posted on Instagram to support his claim that the ball had landed out.}
\label{fig:tennis-intro}
\end{figure}

Disputes over whether technology presumed to be more accurate should take precedence over human judgment are not confined to soccer. In tennis, for instance, electronic line calling has also sparked much controversy. A recent example involved Alexander Zverev, then ranked No. 2 in the world. During his match against Alejandro Davidovich Fokina at the 2025 Madrid Open, Zverev disputed an electronic line call in the second set. He pointed to the ball mark on the clay, arguing that the line-calling technology had made an error of four to five centimeters. Umpire Mohamed Lahyani refused to inspect the ball mark on the court because, under the electronic line-calling rules, the electronic call took precedence. Out of frustration, Zverev retrieved his phone from his bag and took a picture of the mark, later posting it on social media. This incident further fueled the ongoing debate over how to resolve conflicts between human and machine refereeing systems~\cite{zverev}. Figure~\ref{fig:tennis-intro} shows two images associated with this incident.

The tension between human judgment and technology extends beyond sports to various other applications of artificial intelligence (AI). In many such settings, the human provides a binary judgment, while the AI provides a numerical measurement that can be converted into a binary decision by comparison with a threshold. In medical diagnosis, for example, a physician may classify a lesion as benign or malignant, while an AI system assigns it a numerical malignancy score. In aircraft inspection, a maintenance engineer may classify a component as safe or requiring maintenance, while an AI system produces a numerical defect-severity score. In finance, a hedge fund manager may be bullish or bearish on a stock based on experience and judgment, while an AI model produces a numerical estimate of its expected return. These examples share the same information structure: a human binary judgment and an AI numerical measurement concerning the same underlying decision.



In this paper, we develop a maximum-likelihood model for combining human judgment and AI measurements to estimate an unknown quantity of interest and make a final binary decision. We assume that the correct decision is determined by the sign of
a fixed, unknown scalar. The human and AI receive noisy information about this scalar through independent additive Gaussian errors, but only the sign of the human’s internal observation is available. We prove existence and uniqueness of the maximum-likelihood estimate of the unknown scalar and derive an explicit rule for its sign. The resulting decision rule combines the human binary call and the AI numerical measurement and requires only a threshold comparison once the error parameters have been calibrated. We also show how maximum likelihood can be used to estimate these parameters from data.

The model applies to any binary classification problem in which the human and AI receive noisy information about the same underlying scalar quantity and the stated error assumptions hold. We present the model in the
context of offside calls, which provided our original motivation; the
translation to other applications satisfying these assumptions is immediate.
In Section~\ref{sec:prelim}, we review the required probability and optimization
concepts. In Section~\ref{sec:model}, we derive the model and decision rule.
In Section~\ref{sec:examples}, we present numerical examples and illustrations of decision boundaries. In Section~\ref{sec:calibration}, we derive the calibration
procedures, and in Section~\ref{sec:conclusion}, we summarize our conclusions.

\section{Preliminaries for maximum likelihood}\label{sec:prelim}
In this section, we review the probability and optimization concepts needed to develop the maximum-likelihood model in the sections that follow.

\subsection{The Gaussian distribution}
A real-valued random variable $X$ has a \emph{Gaussian distribution} with
mean $m\in\R$ and standard deviation $\sigma>0$, written
$X\sim N(m,\sigma^2)$, if its probability density function is
\begin{equation}\label{eq:normal-density}
 f_X(x)=\frac{1}{\sigma\sqrt{2\pi}}
 \exp\!\left(-\frac{(x-m)^2}{2\sigma^2}\right),\qquad x\in\R.
\end{equation}
The density determines interval probabilities by
$\Prob(a\le X\le b)=\int_a^b f_X(x)\,dx$ for $a\le b$.
Its mean and variance satisfy
\[
 \E[X]=\int_{-\infty}^{\infty}x f_X(x)\,dx=m,\qquad
 \Var(X)=\int_{-\infty}^{\infty}(x-m)^2 f_X(x)\,dx=\sigma^2.
\]
The standard deviation $\sigma$ is the square root of the variance and has the same units as $X$.

We denote the density and cumulative distribution function of $N(0,1)$ as
\begin{equation}\label{eq:normal-functions}
 \phi(u)=\frac{1}{\sqrt{2\pi}}e^{-u^2/2},\qquad
 \Phi(u)=\int_{-\infty}^{u}\phi(t)\,dt.
\end{equation}
The cumulative distribution function of $X\sim N(m,\sigma^2)$ is therefore
\begin{equation}\label{eq:normal-cdf}
 F_X(x):=\Prob(X\le x)=\Phi\!\left(\frac{x-m}{\sigma}\right).
\end{equation}
We will use the identities
$\Phi(-u)=1-\Phi(u)$, $\Phi'(u)=\phi(u)$, and
$\phi'(u)=-u\phi(u)$.

Gaussian distributions are widely used in statistics and probability theory to model additive measurement error. The central limit theorem (see, e.g.,~\cite[Section~5.4]{bertsekas-tsitsiklis}) provides a useful justification for this choice. Roughly speaking, when an error is the sum of many small independent sources of error, its distribution can often be well approximated by a Gaussian distribution. This motivates the error model used in this paper.



\subsection{Likelihood and log-likelihood}
Let $X$ denote data generated by a probability model with an unknown fixed
parameter $\vartheta\in\Theta$, where $\Theta$ is the parameter space. If $X$ is
discrete, let $p_{\vartheta}(x)=\Prob_{\vartheta}(X=x)$ be its probability mass
function. If $X$ has a probability density function, denote that function by
$f_{\vartheta}$. For an observed value $x$, the \emph{likelihood function} is
\[
 L(\vartheta;x)=
 \begin{cases}
 p_{\vartheta}(x),&\text{in the discrete case},\\
 f_{\vartheta}(x),&\text{in the continuous case}.
 \end{cases}
\]
We hold the data $x$ fixed and regard $L(\vartheta;x)$ as a function of
$\vartheta$. When the data points come from independent observations, the joint likelihood is the product
of their probability masses or probability densities, as appropriate for each
observation. A \emph{maximum-likelihood estimate} (see, e.g.,~\cite[Chap. 7]{boyd}) is any parameter value
satisfying
\begin{equation}\label{eq:generic-mle}
 \widehat\vartheta\in\argmax_{\vartheta\in\Theta}L(\vartheta;x).
\end{equation}
Here $\argmax$ denotes the set of maximizers. For discrete data, maximum likelihood selects a parameter value that assigns
the largest probability to the observed data. For continuous data, the
probability of any particular observed value is zero. We instead maximize the
density at that value. When the density is continuous at the observation, the
probability of a small neighborhood is approximately the density there times
the neighborhood's volume. Thus the maximum likelihood approach focuses on the local concentration of probability around the observed data.


Note that the likelihood function is nonnegative. We define the \emph{log-likelihood} function by $\ell(\vartheta;x)=\log L(\vartheta;x)$ when
$L(\vartheta;x)>0$, with the convention $\log 0=-\infty$. Since the logarithm
is strictly increasing on $(0,\infty)$, maximizing the likelihood and
maximizing the log-likelihood are equivalent. Taking logarithms turns products
into sums, which is convenient for some calculations. All likelihoods considered below are finite and strictly positive
at every admissible parameter value.

\subsection{Concavity and maximization}
A set $C\subseteq\R^n$ is \emph{convex} if it
contains the line segment between any two of its points; that is,
\[
 (1-t)x+ty\in C\qquad\text{for every }x,y\in C\text{ and }t\in[0,1].
\]
A function $f:C\to\R$ on a convex set $C$ is \emph{concave} if
\begin{equation}\label{eq:concavity}
 f((1-t)x+ty)\ge (1-t)f(x)+tf(y)
 \quad\text{for all }x,y\in C\text{ and }t\in[0,1].
\end{equation}
It is \emph{strictly concave} if the inequality is strict whenever $x\ne y$
and $t\in(0,1)$.

\begin{proposition}\label{prop:unique}
A strictly concave function on a convex set has at most one maximizer.
\end{proposition}
\begin{proof}
If distinct points $x$ and $y$ both attained the maximum value $M$, strict
concavity at their midpoint would give
$f((x+y)/2)>(f(x)+f(y))/2=M$, a contradiction.
\end{proof}

Let $f$ be twice continuously differentiable on an open interval $I$. If
$f''(x)<0$ for every $x\in I$, then $f$ is strictly concave
\cite[Section~3.1.4]{boyd}, and $f'$ is strictly decreasing.
Consequently, if $f'(x^\star)=0$ at some $x^\star\in I$, then $f'>0$ to its
left and $f'<0$ to its right, so $x^\star$ is the unique maximizer on $I$.
Note that a strictly concave function does not necessarily have a maximizer (consider, e.g., $f(x)=-e^{-x}$ for $x\in\mathbb{R}$). The claim is that if it has one, then the maximizer must be unique.

\section{Combining the referee's call and the AI measurement}\label{sec:model}
In this section, we formulate a model for combining a human binary call with
an AI numerical measurement and derive the corresponding maximum-likelihood
decision. The model applies to any binary classification problem in which the two observations satisfy the Gaussian error assumptions stated below. We present the derivation in the context of offside calls as a concrete example.

\subsection{The observation model}
For an attacker in the opponents' half, let $\theta\in\R$ be the fixed,
unknown \emph{true margin}: the signed longitudinal distance of the attacker's
foremost eligible body part beyond the offside line at the relevant teammate
contact. We use ``offside'' and ``onside'' for this positional classification.
Positive margins indicate offside; nonpositive margins indicate onside.
Define the binary decision function
\begin{equation}\label{eq:true-label}
 T(t)=\begin{cases}+1,&t>0,\\-1,&t\le0,\end{cases}
 \qquad t\in\R.
\end{equation}
The correct decision is $T(\theta)$; the challenge is that the exact value of
$\theta$ is unknown.


We model the referee's perception of the true margin by an unobserved random variable $X_{\HH}$. Specifically, $X_{\HH}$ equals the true margin $\theta$ plus Gaussian noise with mean $\mH$ and standard deviation $\sdH$. The referee does not explicitly compute a numerical distance; rather, $X_{\HH}$ represents the referee's latent internal assessment in our model. The observed \emph{human call} is only the binary decision $H=T(X_{\HH})$. Specifically,
\begin{equation}\label{eq:human-model}
 X_{\HH}=\theta+\varepsilon_{\HH},\qquad
 \varepsilon_{\HH}\sim N(\mH,\sdH^2),\qquad H=T(X_{\HH}),
\end{equation}
and only $H$, i.e. the referee offside/onside call, is observed.

The AI measurement is also modeled based on additive Gaussian noise, but with a possibly different mean and standard deviation:
\begin{equation}\label{eq:ai-model}
 Y=\theta+\varepsilon_{\AI},\qquad
 \varepsilon_{\AI}\sim N(\mA,\sdA^2).
\end{equation}
We assume that the human measurement error $\varepsilon_{\HH}$ and the AI
measurement error $\varepsilon_{\AI}$ are independent.\footnote{This assumption is reasonable when the two errors arise primarily from distinct sources, such as human visual perception and the AI system's sensing, modelling, and computation.} We also assume that
$\mH,\mA\in\R$ and $\sdH,\sdA>0$ are known. Section~\ref{sec:calibration}
explains how these parameters can be estimated from data. The means represent bias, and the
standard deviations represent the variability of the measurement errors.
The means and standard deviations have the same distance units as $\theta$.

We observe $H=h\in\{-1,+1\}$ and $Y=y\in\R$, but not $X_{\HH}$.
We refer to $Y$ as the \emph{AI measurement} and to $y$ as its observed value.\footnote{Beyond the soccer example, many AI-based binary classifiers first produce a real-valued score and then compare that score with a threshold to make a binary decision. Examples include the output score of a neural network or the signed distance to the separating hyperplane in a support vector machine. In such cases, the underlying numerical score is available.}
Define the \emph{bias-corrected AI measurement} by
\begin{equation}\label{eq:z}
 Z=Y-\mA,\qquad z=y-\mA.
\end{equation}
Then $Z\sim N(\theta,\sdA^2)$, and $z$ is the observed bias-corrected AI measurement.

\subsection{The joint likelihood}
\begin{figure}[!htbp]
\centering\includegraphics[width=.80\linewidth]{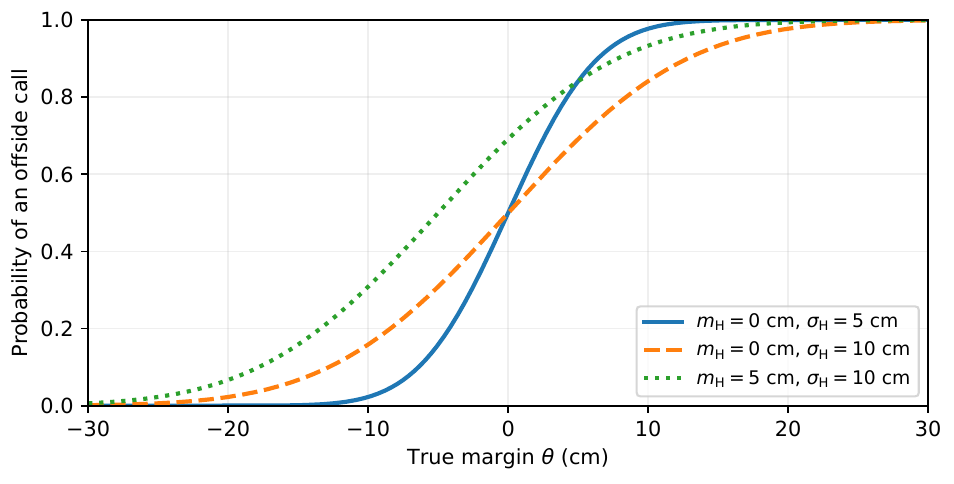}
\caption{Human offside-call probabilities from~\eqref{eq:binary-mass}.
A larger $\sdH$ makes the transition more gradual. A nonzero \(m_H\) shifts the 50\% crossover point from \(\theta=0\) to \(\theta=-m_H\). The parameters are illustrative.}
\label{fig:response}
\end{figure}

We first derive the probability mass function of the human call $H$. Recall that $\theta$ is the fixed true margin. For any event $A$ involving the random human call or AI measurement, let $\Prob_\theta(A)$ denote the probability that $A$ occurs under the model when the true margin is $\theta$. For example, \(\Prob_\theta(H=+1)\) is the probability that the referee calls offside when the true margin is \(\theta\).
From~\eqref{eq:human-model}, an offside call occurs when
$\varepsilon_{\HH}>-\theta$. The symmetry of the Gaussian distribution gives
\[
 \Prob_\theta(H=+1)
 =1-\Phi\!\left(\frac{-\theta-\mH}{\sdH}\right)
 =\Phi\!\left(\frac{\theta+\mH}{\sdH}\right).
\]
The complementary probability gives the onside case. Thus, for
$h\in\{-1,+1\}$,
\begin{equation}\label{eq:binary-mass}
 \Prob_\theta(H=h)=\Phi\!\left(\frac{h(\theta+\mH)}{\sdH}\right).
\end{equation}
In Figure~\ref{fig:response}, we plot the offside-call probability from
\eqref{eq:binary-mass} for several human error parameters. Increasing
$\sdH$ makes the transition between predominantly onside and predominantly
offside calls more gradual. A nonzero \(m_H\) shifts the true-margin value at which the referee is equally likely to call onside or offside from \(\theta=0\) to \(\theta=-m_H\).


We next consider the AI measurement. By \eqref{eq:ai-model}, $Y$ has a Gaussian distribution with mean $\theta+\mA$ and standard deviation $\sdA$. In view of (\ref{eq:normal-density}) and (\ref{eq:normal-functions}), the probability density function of $Y$, evaluated at the observed value $y$, is given by
\[
f_Y(y)
=
\frac{1}{\sigma_{\mathrm{AI}}}
\phi\!\left(
\frac{y-\theta-m_{\mathrm{AI}}}{\sigma_{\mathrm{AI}}}
\right).
\]


Because the measurement errors $\varepsilon_{\HH}$ and $\varepsilon_{\AI}$ are independent, the human call $H=T(\theta+\varepsilon_{\HH})$ and the AI measurement $Y=\theta+\varepsilon_{\AI}$ are independent for each fixed $\theta$. Multiplying the probability mass of $H=h$ by the probability density of $Y$ at $y$ therefore gives the joint likelihood \begin{equation}\label{eq:likelihood}
 L(\theta;h,y)=
 \Phi\!\left(\frac{h(\theta+\mH)}{\sdH}\right)
 \frac{1}{\sdA}\phi\!\left(\frac{y-\mA-\theta}{\sdA}\right).
\end{equation}
Here $\phi$ and $\Phi$ are the standard Gaussian probability density function
and cumulative distribution function defined in \eqref{eq:normal-functions}.
The human factor uses $\Phi$ because only a sign is observed: we integrate the
density of $X_{\HH}$ over the half-line corresponding to the reported call.
The AI factor uses $\phi$ because the numerical value $y$ itself is observed.
Thus the first factor is a probability mass and the second is a probability
density. In particular, for real $a\le b$,
\[
 \Prob_\theta(H=h,\ a\le Y\le b)=\int_a^b L(\theta;h,y)\,dy.
\]

\subsection{The maximum-likelihood estimate}
We now maximize the joint likelihood $L(\theta;h,y)$ in \eqref{eq:likelihood}
over the true margin $\theta\in\R$, holding the observed human call $h$ and
AI measurement $y$ fixed. Recall from \eqref{eq:z} that
$z=y-\mA$ is the observed bias-corrected AI measurement. By taking the logarithm of the likelihood function in (\ref{eq:likelihood}), we can write down the log-likelihood function as
\begin{equation}\label{eq:loglik}
 \ell(\theta;h,y)=
 -\log(\sdA\sqrt{2\pi})-\frac{(z-\theta)^2}{2\sdA^2}
 +\log\Phi\!\left(\frac{h(\theta+\mH)}{\sdH}\right).
\end{equation}
Our goal is to show that this function attains a unique maximum and to
characterize its maximizer. We first establish a strict-concavity property of the Gaussian cumulative distribution function that will be useful here and again in the calibration analysis of Section~\ref{sec:calibration}.
With $\phi(\cdot)$ and $\Phi(\cdot)$ as in (\ref{eq:normal-functions}), define
\begin{equation}\label{eq:lambda}
 \lambda(u)=\frac{\phi(u)}{\Phi(u)},\qquad u\in\R.
\end{equation}
This function is positive and finite because $\Phi(u)\in(0,1)$ for every $u\in\R$.
The following lemma is standard; we include a proof for completeness.

\begin{lemma}\label{lem:logphi}
Let $\Phi$ be the cumulative distribution function in \eqref{eq:normal-functions}.
The function $u\mapsto\log\Phi(u)$ is strictly concave on $\R$.
\end{lemma}


\begin{proof}
Let $g(u)=\log\Phi(u)$. To prove that $g$ is strictly concave, it suffices to show that $g''(u)<0$ for every $u\in\mathbb{R}$. Using the fact that $\Phi'(u)=\phi(u)$ and the definition of $\lambda$ in~\eqref{eq:lambda}, we have $g'(u)=\frac{\phi(u)}{\Phi(u)}=\lambda(u).$ Hence, $g''(u)=\lambda'(u).$ Since $\phi'(u)=-u\phi(u)$,
\[
\lambda'(u)
=
\frac{\phi'(u)\Phi(u)-\phi(u)\Phi'(u)}{\Phi(u)^2}
=
-\lambda(u)\bigl(u+\lambda(u)\bigr).
\]
We have $\lambda(u)>0$, so it remains to show that $u+\lambda(u)>0$. Since $\phi'(t)=-t\phi(t)$ and $\phi(t)\to0$ as $t\to-\infty$, we have
\[
u\Phi(u)+\phi(u)
=
\int_{-\infty}^{u}(u-t)\phi(t)\,dt
>0.
\]
As $\Phi(u)>0$, division by $\Phi(u)$ gives $u+\lambda(u)>0.$
\end{proof}

\Needspace{8\baselineskip}
\begin{theorem}[Existence and uniqueness of the maximum-likelihood estimate]\label{thm:estimate}
Consider the observation model \eqref{eq:human-model}--\eqref{eq:ai-model}
with independent measurement errors $\varepsilon_{\HH}$ and
$\varepsilon_{\AI}$, given means $\mH,\mA\in\R$, and standard deviations
$\sdH,\sdA>0$. For every observed pair $(h,y)\in\{-1,+1\}\times\R$,
there is a unique maximum-likelihood estimate $\widehat\theta_{\ML}$ of the
true margin $\theta$. It is the unique solution of the equation
\begin{equation}\label{eq:mle-equation}
 \frac{y-\mA-\widehat\theta_{\ML}}{\sdA^2}
 +\frac{h}{\sdH}
 \lambda\!\left(\frac{h(\widehat\theta_{\ML}+\mH)}{\sdH}\right)=0,
\end{equation}
where $\lambda(\cdot)$ is as in (\ref{eq:lambda}).
\end{theorem}
\begin{proof}
Let $z=y-\mA$. For fixed observations $(h,y)$, differentiation of the log-likelihood function with respect to $\theta$ gives
\begin{align}
 \ell'(\theta;h,y)
 &=\frac{z-\theta}{\sdA^2}
   +\frac{h}{\sdH}\lambda\!\left(\frac{h(\theta+\mH)}{\sdH}\right),
   \label{eq:score}\\
 \ell''(\theta;h,y)
 &=-\frac{1}{\sdA^2}
   +\frac{1}{\sdH^2}\lambda'\!\left(\frac{h(\theta+\mH)}{\sdH}\right)
   <0.\label{eq:second-derivative}
\end{align}
The expression for the second derivative uses the fact that $h^2=1$, and the last inequality follows from the proof of Lemma~\ref{lem:logphi} keeping in mind that \(\lambda'(u)=(\log\Phi)''(u)\). We can therefore conclude from (\ref{eq:second-derivative}) that the log-likelihood function is strictly concave.


To prove existence of a maximizer, observe that since $\log\Phi\le0$, we have
\[
 \ell(\theta;h,y)
 \le -\log(\sdA\sqrt{2\pi})-\frac{(z-\theta)^2}{2\sdA^2}
 \longrightarrow-\infty\qquad\text{as }|\theta|\to\infty.
\]
Choose a closed bounded interval containing zero outside which
$\ell(\theta;h,y)<\ell(0;h,y)$. Since $\ell$ is a continuous function, by the extreme value theorem, it must have a maximizer on this interval,
which is also a maximizer on $\R$.
Strict concavity of $\ell$ and Proposition~\ref{prop:unique} imply that this maximizer must be unique. Lastly, the maximizer is the solution to $\ell'=0$, which in view of (\ref{eq:score}) gives the equation in (\ref{eq:mle-equation}). 
\end{proof}

Rearranging~\eqref{eq:mle-equation}, we have
\begin{equation}\label{eq:adjustment}
 \widehat\theta_{\ML}=z+
 \frac{h\sdA^2}{\sdH}
 \lambda\!\left(\frac{h(\widehat\theta_{\ML}+\mH)}{\sdH}\right).
\end{equation}
Since $\lambda$ is positive (see \eqref{eq:lambda}),
an offside call increases the maximum likelihood estimate relative to the bias-corrected AI measurement $z$, and an onside call decreases it. The quantity $\widehat\theta_{\ML}$ can be computed, e.g., by bisection~\cite[Lecture 7]{AAA-363} on the strictly
decreasing derivative~\eqref{eq:score}. Note that \(\ell'(\theta)\to+\infty\) as \(\theta\to-\infty\) and \(\ell'(\theta)\to-\infty\) as \(\theta\to+\infty\), so two points with opposite derivative signs can always be found to serve as the initial bracket for the bisection algorithm.

\begin{figure}[!htbp]
\centering\includegraphics[width=.77\linewidth]{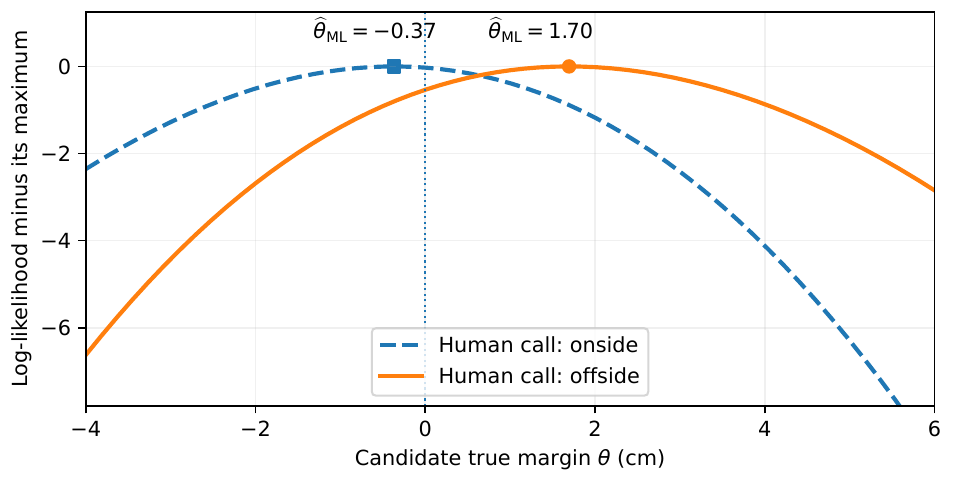}
\caption{Log-likelihoods for $y=1$~cm, $\mH=\mA=0$, and
$\sdH=\sdA=2$~cm. Each curve has been shifted to have maximum zero.
Changing the human call changes the sign of $\widehat\theta_{\ML}$ even
though the observed AI measurement is unchanged.}
\label{fig:likelihood}
\end{figure}

In Figure~\ref{fig:likelihood}, we fix the observed AI measurement $y$ and the four error
parameters and plot the log-likelihood for each human call. In this example,
changing the human call changes the sign of the unique maximizer. In the two cases, the maximum-likelihood estimate lies to the right or left of the observed AI measurement by unequal amounts. The intuition is that an agreeing human call provides relatively little additional information because such a call was already likely at values near the AI measurement. By contrast, a disagreeing human call is more informative because it is less likely under those same values, causing the likelihood to shift farther to accommodate it.
\FloatBarrier

\subsection{The maximum-likelihood decision}
Recall that $\widehat\theta_{\ML}$ is the unique maximum-likelihood estimate
of the true margin $\theta$, and is characterized by \eqref{eq:mle-equation}. We define the
\emph{maximum-likelihood decision} to be $T(\widehat\theta_{\ML})$, where
$T$ is the binary decision function in \eqref{eq:true-label}. This will be our final decision obtained by combining the referee's call with the AI measurement: our call will be offside if $T(\widehat\theta_{\ML})=1$, and onside if $T(\widehat\theta_{\ML})=-1$.  We next show that
this decision can be obtained without computing the numerical value of
$\widehat\theta_{\ML}$.

In the following proposition, $h\in\{-1,1\}$ is the observed human call, $y\in\mathbb{R}$ is the observed
AI measurement, and $\mH,\mA,\sdH,\sdA$ are the error parameters in
\eqref{eq:human-model}--\eqref{eq:ai-model}. The functions $\phi(\cdot)$ and $\Phi(\cdot)$
are defined in \eqref{eq:normal-functions}.

\begin{proposition}[Maximum-likelihood decision rule]\label{prop:decision}
Under the assumptions of Theorem~\ref{thm:estimate}, the maximum-likelihood
decision is offside if and only if
\begin{equation}\label{eq:general-decision}
 y-\mA+
 \frac{h\sdA^2}{\sdH}
 \frac{\phi(h\mH/\sdH)}{\Phi(h\mH/\sdH)}>0.
\end{equation}
\end{proposition}
\begin{proof}
The maximum-likelihood decision is offside if $\widehat\theta_{\ML}>0$ and onside if $\widehat\theta_{\ML}
\leq 0$. For the log-likelihood $\ell(\theta;h,y)$ in \eqref{eq:loglik}, the derivative
with respect to $\theta$ is given by \eqref{eq:score}. From the proof of Theorem~\ref{thm:estimate}, we see that $ \ell'$ is strictly decreasing and vanishes at $\widehat\theta_{\ML}$. Hence,
\[
 \widehat\theta_{\ML}>0
 \quad\Longleftrightarrow\quad \ell'(0;h,y)>0.
\]
Evaluating~\eqref{eq:score} at zero and multiplying by $\sdA^2$ gives
\eqref{eq:general-decision}. 
\end{proof}


For \(h\in\{-1,+1\}\), define the positive, referee-call-dependent threshold
\begin{equation}\label{eq:thresholds}
 c_h=\frac{\sdA^2}{\sdH}\lambda\!\left(\frac{h\mH}{\sdH}\right),
\end{equation}
where $\lambda(\cdot)$ is as in \eqref{eq:lambda}.
Recall that \(z=y-m_{\mathrm{AI}}\) is the observed bias-corrected AI measurement.
The maximum-likelihood decision rule declares offside when $z+h c_h>0.$  Thus, following a human offside call, the final decision is offside if \(z>-c_{+1}\), whereas following a human onside call, the final decision is offside if \(z>c_{-1}\). The two thresholds $c_{+1}$ and $c_{-1}$ can be computed offline. 

As a sanity check, consider the case where \(m_H>0\), which corresponds to a referee who may be somewhat biased to call offside. Then, $\frac{m_H}{\sigma_H}>-\frac{m_H}{\sigma_H}$. Since \(\lambda\) is strictly decreasing (see the proof of Lemma~\ref{lem:logphi}), we have $\lambda\!\left(\frac{m_H}{\sigma_H}\right)
<\lambda\!\left(-\frac{m_H}{\sigma_H}\right)$,
and therefore \(c_{+1}<c_{-1}\). Hence, when the referee is biased toward offside, the AI evidence required to overturn a human onside call in favor of offside is stronger than the AI evidence required to overturn a human offside call in favor of onside.



\subsection{The special case of zero means}\label{sec:zero}
We now specialize the referee-call-dependent thresholds $c_{+1}$ and $c_{-1}$ from
\eqref{eq:thresholds} to the prominent case where $\mH=\mA=0$. The observed bias-corrected AI
measurement $z=y-\mA$ then equals the observed AI measurement $y$.
Since $\lambda(0)=\phi(0)/\Phi(0)=\sqrt{2/\pi}$, the two thresholds reduce to
\begin{equation}\label{eq:cml}
 c_{\ML}=\sqrt{\frac{2}{\pi}}\frac{\sdA^2}{\sdH}.
\end{equation}
Therefore, the maximum-likelihood decision is offside if and only if
\begin{equation}\label{eq:zero-decision}
 z+h c_{\ML}>0.
\end{equation}
Note that both $z$ and $c_{\ML}$ have units of distance. If $hz\geq 0$, the human call is retained. If \(hz<0\), the human call and AI measurement disagree, and we proceed as follows:
\begin{equation}\label{eq:overturn}
 \begin{aligned}
 |z|>c_{\ML}&\quad\Longrightarrow\quad\text{overturn the human call},\\
 |z|<c_{\ML}&\quad\Longrightarrow\quad\text{retain the human call}.
 \end{aligned}
\end{equation}
At $|z|=c_{\ML}$ in a disagreement, the final decision is onside. This
retains an onside call and overturns an offside call.


It is intuitive that increasing \(\sigma_{\mathrm{AI}}\) raises the threshold \(c_{\mathrm{ML}}\), since a noisier AI requires stronger evidence to overturn the human call, whereas increasing \(\sigma_H\) lowers \(c_{\mathrm{ML}}\) because a noisier human call should be easier to overturn.
The different powers of $\sigma_{\mathrm{AI}}$ and $\sigma_H$ in $c_{\mathrm{ML}}$ come from the two likelihood terms: differentiating the Gaussian log-density for the AI measurement introduces a factor $1/\sigma_{\mathrm{AI}}^2$, whereas differentiating the log Gaussian tail probability for the human call introduces a factor $1/\sigma_H$, as seen in \eqref{eq:score}.


\paragraph{Decision error probabilities.}
Recall the definition of the binary decision function $T(\cdot)$ from~(\ref{eq:true-label}). We now compare, under our model, the probabilities that the maximum-likelihood decision rule $T(\widehat\theta_{\ML})$, the human call $H$, and the AI decision $T(Y)$ make an incorrect decision, i.e., a decision different from $T(\theta)$. 

Let us continue with the case $\mH=\mA=0$. For a fixed
true margin $\theta>0$, an erroneous maximum-likelihood onside decision occurs when $H=+1$ and
$Z\le-c_{\ML}$, or when $H=-1$ and $Z\le c_{\ML}$. Here, $Z$ (which equals $Y$ since $\mA=0$) is the
bias-corrected AI measurement in \eqref{eq:z} and $c_{\ML}$ is the threshold
in \eqref{eq:cml}. Independence of the human and AI measurement errors, \(\epsilon_H\) and \(\epsilon_{\mathrm{AI}}\), gives
\[
\begin{aligned}
 \Prob_\theta\bigl(T(\widehat\theta_{\ML})\ne T(\theta)\bigr)
 &=\Phi\!\left(\frac{\theta}{\sdH}\right)
   \Phi\!\left(\frac{-\theta-c_{\ML}}{\sdA}\right)\\
 &\quad+\Phi\!\left(-\frac{\theta}{\sdH}\right)
   \Phi\!\left(\frac{c_{\ML}-\theta}{\sdA}\right).
\end{aligned}
\]
For $\theta<0$, by symmetry, we get the same expression with $\theta$ replaced by
$|\theta|$. Lastly, at $\theta=0$, the error probability is $1/2$. 

\begin{figure}[!htbp]
\centering\includegraphics[width=.80\linewidth]{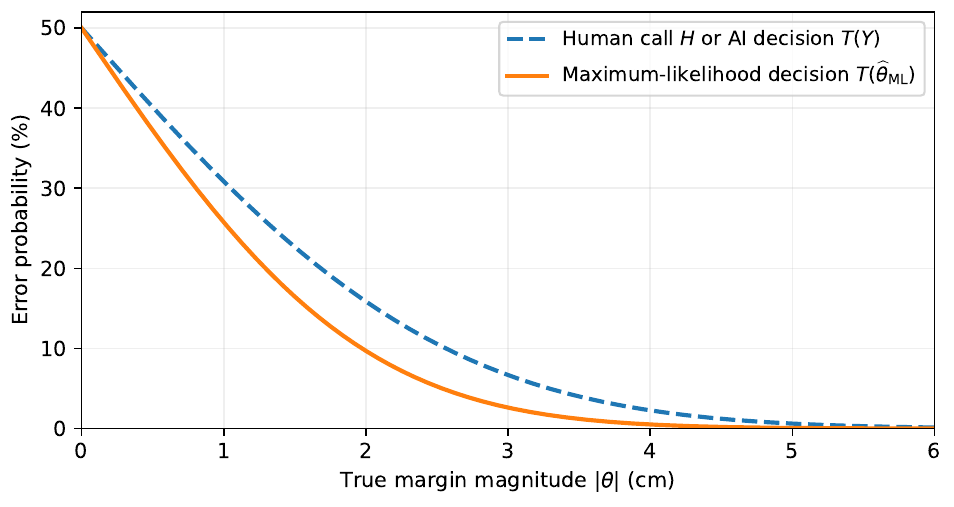}
\caption{Decision error probabilities for $\mH=\mA=0$ and $\sdH=\sdA=2$~cm.
The human call $H$ and the AI decision $T(Y)$ have the same error probability.}
\label{fig:error}
\end{figure}

The human call and AI decision separately have error probabilities $\Phi(-|\theta|/\sdH)$
and $\Phi(-|\theta|/\sdA)$. Figure~\ref{fig:error} compares these probabilities
when $\sdH=\sdA=2$~cm. At $|\theta|=2$~cm, the maximum-likelihood decision
has error probability approximately $9.70\%$, compared with $15.87\%$ for
either source alone. These probabilities average over all observations at a
fixed true margin, not only disagreements. 


\section{Numerical examples and decision boundaries}\label{sec:examples}
In this section, we illustrate the maximum-likelihood decision rule with a few numerical examples and describe the boundaries between retaining and overturning the human call.


\subsection{Examples with zero and nonzero means}
\paragraph{Example 1 (Zero means).}
Recall the zero-mean threshold $c_{\ML}=\sqrt{2/\pi}\,\sdA^2/\sdH$ from
\eqref{eq:cml}. Let $\mH=\mA=0$, $\sdH=10$~cm, and $\sdA=2$~cm.
Then $c_{\ML}\approx0.3192$~cm. The observed bias-corrected AI measurement
$z=y-\mA$ equals $y$ in this case. Table~\ref{tab:examples} shows that a
conflicting measurement of magnitude $0.20$~cm retains the human call,
whereas a magnitude of $0.50$~cm overturns it. The estimates are computed from
\eqref{eq:mle-equation}; their signs follow directly from
\eqref{eq:zero-decision}.

\begin{table}[!htbp]
\centering
\caption{Maximum-likelihood decisions for $\mH=\mA=0$, $\sdH=10$~cm,
and $\sdA=2$~cm. Numerical estimates are rounded to four decimal places.}
\label{tab:examples}
\begin{tabular}{@{}lrrll@{}}
\toprule
Human call & $z$ (cm) & $\widehat\theta_{\ML}$ (cm) & Final decision & Action\\
\midrule
Offside & $-0.20$ & $ 0.1162$ & Offside & Retain\\
Offside & $-0.50$ & $-0.1763$ & Onside & Overturn\\
Onside & $ 0.20$ & $-0.1162$ & Onside & Retain\\
Onside & $ 0.50$ & $ 0.1763$ & Offside & Overturn\\
\bottomrule
\end{tabular}
\end{table}

\FloatBarrier
\paragraph{Example 2 (Unequal means).}
We next show how unequal error means affect the maximum-likelihood decision. In the human model,
a negative $\mH$ can represent a reluctance to call offside in a tense
situation, whereas a positive $\mH$ represents a tendency to call offside.
Keep $\sdH=10$~cm and $\sdA=2$~cm, but set $\mH=5$~cm and $\mA=0$.
The thresholds~\eqref{eq:thresholds} become
\[
 c_{+1}\approx0.2037\text{ cm},\qquad
 c_{-1}\approx0.4564\text{ cm}.
\]
An offside call with $y=-0.30$~cm is overturned, but an onside call with
$y=0.30$~cm is retained. Equal conflicting magnitudes have different effects
because the human bias makes the thresholds asymmetric.


To isolate the effect of AI bias, instead set \(m_H=0\) and \(m_{\mathrm{AI}}=1\) cm, keeping the same standard deviations. Since \(m_H=0\), the two thresholds in \eqref{eq:thresholds} are equal to \(c_{\mathrm{ML}}\) in \eqref{eq:cml}. For \(h=+1\) and \(y=0.50\) cm,
\[
z+c_{\mathrm{ML}}
=y-m_{\mathrm{AI}}+c_{\mathrm{ML}}
\approx -0.1808\text{ cm}<0.
\]
The maximum-likelihood decision is onside. Omitting the known AI bias would
produce an offside decision for the same observations.

\subsection{Decision boundaries}
\begin{figure}[!htbp]
\centering\includegraphics[width=.82\linewidth]{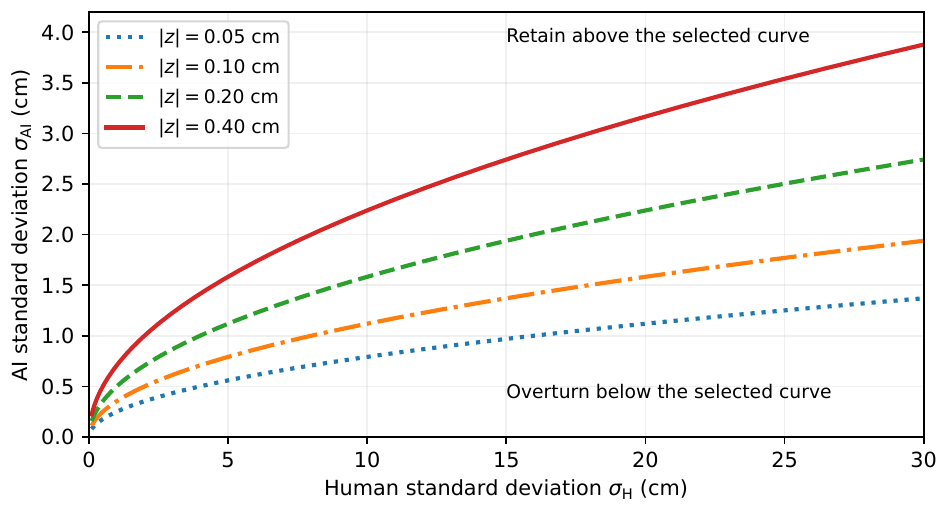}
\caption{Decision boundaries for $|z|=0.05$, $0.10$, $0.20$, and $0.40$~cm,
with both means zero and $hz<0$. For each curve, we overturn the human decision below the curve and retain it above.}
\label{fig:curves}
\end{figure}

We next visualize the boundary between retaining and overturning the human call. Suppose that both means are zero and that the observed bias-corrected AI
measurement $z=y$ conflicts with the human call, so $hz<0$. Let $c_{\ML}$ be the threshold in (\ref{eq:cml}). For a fixed
magnitude $|z|>0$, the boundary $|z|=c_{\ML}$ can be written as
\begin{equation}\label{eq:boundary}
 \sdA=\left(\frac{\pi}{2}\right)^{1/4}\sqrt{|z|\sdH}.
\end{equation}
Figure~\ref{fig:curves} plots this boundary in the $(\sdH,\sdA)$ plane for
four magnitudes of the conflicting AI measurement: $|z|=0.05$, $0.10$,
$0.20$, and $0.40$~cm. For a given
curve, a point below it satisfies $|z|>c_{\ML}$ and therefore overturns the
human call. A point above it retains the call. Increasing $|z|$ raises the
boundary and enlarges the overturning region.

\begin{figure}[!htbp]
\centering\includegraphics[width=.80\linewidth]{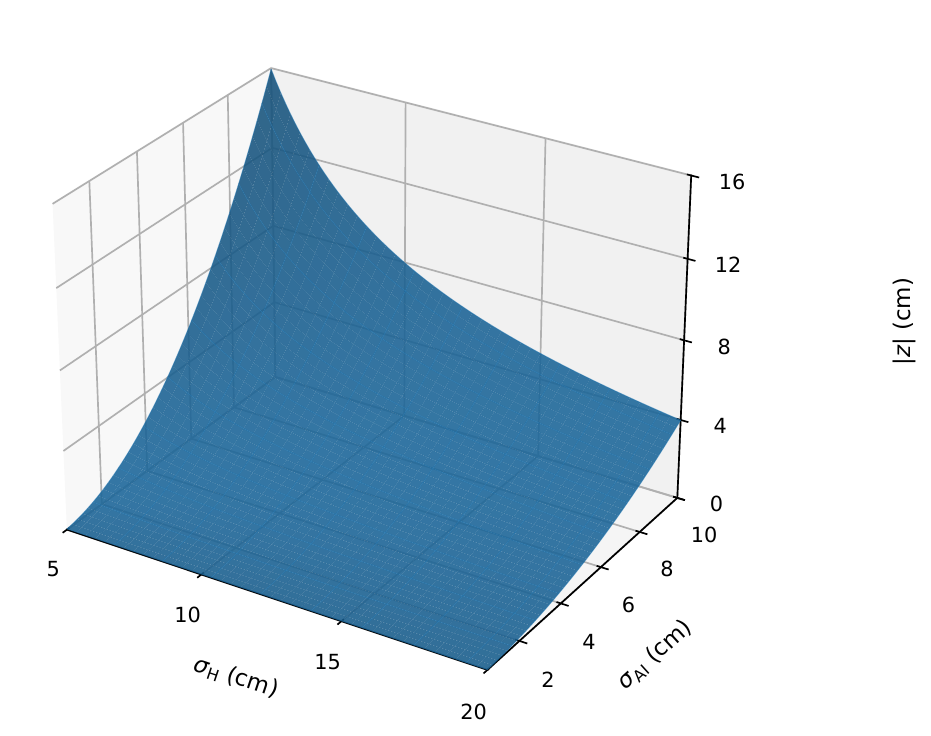}
\caption{The decision boundary in~\eqref{eq:surface}. Both means are zero, and $hz<0$.
Above the surface, we overturn the human decision; below it, we retain it. On the surface, the estimated true margin is zero and the final decision is onside.}
\label{fig:surface}
\end{figure}

Allowing the conflicting AI measurement magnitude $|z|$ to vary gives the surface
\begin{equation}\label{eq:surface}
 |z|=\sqrt{\frac{2}{\pi}}\frac{\sdA^2}{\sdH}
\end{equation}
in Figure~\ref{fig:surface}. Points above the surface overturn the human call;
points below retain it. A horizontal slice at a fixed $|z|$ gives the
corresponding two-dimensional boundary~\eqref{eq:boundary}.


\FloatBarrier

\section{Maximum-likelihood calibration}\label{sec:calibration}

The observation model \eqref{eq:human-model}--\eqref{eq:ai-model}  depends on four parameters: the human error mean $\mH$ and standard deviation $\sdH$, and the AI error mean $\mA$ and standard deviation $\sdA$. In this section, we show how to estimate these parameters from calibration data. The same maximum-likelihood framework used above can also be used for this purpose once reference margins are available.



For our running offside example, calibration data could be obtained from controlled plays or reconstructed incidents for which the player positions and contact times are known. The human calibration sample would record the referee's binary call at each known true margin, while the AI calibration sample would record the corresponding numerical measurement. The two procedures need not use the same observations or the same sample size. More generally, depending on the application, the true margin may become observable after the relevant outcome is realized, or it may be measured during calibration using a higher-fidelity and more costly reference technology that need not be deployed in routine operation.

The first situation arises naturally in finance. For example, a hedge fund manager may predict whether a stock will outperform or underperform a benchmark over a specified horizon, while an AI model produces a numerical prediction of the stock's excess return. Once the horizon has elapsed, the realized excess return provides the true margin: its sign determines the correct binary decision, while its magnitude measures the extent of outperformance or underperformance. Repeating this process across stocks and time periods provides calibration data for estimating the human and AI error parameters.


\subsection{AI parameters}
We collect $n\ge2$ calibration records
\begin{equation}\label{eq:calibration-data}
 (\theta_i,y_i),\qquad i=1,\ldots,n,
\end{equation}
where $\theta_i$ is a fixed, known reference margin and $y_i$ is the observed
AI measurement. We assume that the AI measurement errors \(y_i-\theta_i\), \(i=1,\ldots,n\), are independent and identically distributed according to \(N(m_{\mathrm{AI}},\sigma_{\mathrm{AI}}^2)\). The following derivation is standard; we include it for completeness.

Define the observed AI residuals by \(e_i=y_i-\theta_i\), for \(i=1,\ldots,n\). The likelihood of the observed residual sample, as a function of \(m_{\mathrm{AI}}\) and \(\sigma_{\mathrm{AI}}\), and its log-likelihood are
\begin{align}
 L_{\AI}(\mA,\sdA)
 &=\prod_{i=1}^{n}\frac{1}{\sdA}
   \phi\!\left(\frac{e_i-\mA}{\sdA}\right),\label{eq:ai-calibration-lik}\\
 \ell_{\AI}(\mA,\sdA)
 &=-n\log\sdA-\frac n2\log(2\pi)
   -\frac{1}{2\sdA^2}\sum_{i=1}^{n}(e_i-\mA)^2.
   \label{eq:ai-calibration-log}
\end{align}
Our goal is to find values for $\mA\in\R$ and $\sdA>0$ that maximize $\ell_{\AI}$ (or equivalently $ L_{\AI}$).
Let $\overline e=\frac{1}{n}\sum_{i=1}^{n}e_i$. The identity
\[
 \sum_{i=1}^{n}(e_i-\mA)^2
 =\sum_{i=1}^{n}(e_i-\overline e)^2+n(\mA-\overline e)^2
\]
shows that, for every fixed $\sdA>0$, the unique maximizing mean is
$\mA=\overline e$. Let
$S=\sum_{i=1}^{n}(e_i-\overline e)^2$, and note that $S>0$ when not all sample errors are identical. It is straightforward to verify that
\begin{equation}\label{eq:ai-sd-derivative}
 \frac{d}{d\sdA}\ell_{\AI}(\overline e,\sdA)
 =\frac{S-n\sdA^2}{\sdA^3}.
\end{equation}
This derivative is positive for $\sdA<\sqrt{S/n}$ and negative
for $\sdA>\sqrt{S/n}$. Hence, with $S>0$, the unique maximum-likelihood estimates are the following intuitive quantities:
\begin{equation}\label{eq:ai-calibration}
 \widehat m_{\AI}=\overline e,\qquad
 \widehat\sigma_{\AI}
 =\left(\frac1n\sum_{i=1}^{n}(e_i-\overline e)^2\right)^{1/2}.
\end{equation}
The estimated mean is the average difference between the AI measurement and
the reference margin, so it measures the AI's systematic bias. The estimated
standard deviation is the root mean squared deviation of these residuals from
their average, so it measures their variability after removing the estimated
bias.


\subsection{Human parameters}


We collect $n\ge2$ calibration records $(\theta_i,h_i)$, for
$i=1,\ldots,n$, where $\theta_i$ is a fixed, known reference margin and
$h_i\in\{-1,+1\}$ is the observed human call. For reliable estimation, the calibration sample should include reference margins near the transition region, where both onside and offside calls occur. We assume that the underlying human measurement errors in these records are independent and identically distributed according to \(N(m_H,\sigma_H^2)\). The internal human measurements $X_{\HH,i}$ are unobserved.
By \eqref{eq:binary-mass}, the likelihood of the observed human calls is
\begin{equation}\label{eq:human-calibration-lik}
 L_{\HH}(\mH,\sdH)=
 \prod_{i=1}^{n}\Phi\!\left(\frac{h_i(\theta_i+\mH)}{\sdH}\right).
\end{equation}
The maximum-likelihood problem of interest is therefore
\begin{equation}\label{eq:human-calibration-log}
 \underset{\mH\in\R,\ \sdH>0}{\operatorname{maximize}}\quad
 \ell_{\HH}(\mH,\sdH)
 :=\sum_{i=1}^{n}\log\Phi\!\left(
       \frac{h_i(\theta_i+\mH)}{\sdH}\right).
\end{equation}
It is convenient to introduce the one-to-one reparameterization:
\begin{equation}\label{eq:probit-parameters}
 \alpha=\frac{\mH}{\sdH},\qquad
 \beta=\frac1{\sdH}>0,
 \qquad
 \mH=\frac{\alpha}{\beta},\quad \sdH=\frac1\beta.
\end{equation}
The equivalent problem becomes
\begin{equation}\label{eq:probit}
 \underset{\alpha\in\R,\ \beta>0}{\operatorname{maximize}}\quad
 \widetilde\ell_{\HH}(\alpha,\beta)
 :=\sum_{i=1}^{n}\log\Phi\bigl(h_i(\alpha+\beta\theta_i)\bigr).
\end{equation}
In this parameterization, the probability of an offside call at the known
reference margin $\theta_i$ is $\Phi(\alpha+\beta\theta_i)$. A binary-response
model of this form is called \emph{probit regression}. Problem
\eqref{eq:probit} estimates its two coefficients by maximum likelihood, with
$\beta>0$ required by the fact that $\sdH>0$.

Unlike the AI calibration problem, the human calibration problem does not generally admit closed-form maximum-likelihood estimates. Nevertheless, its optimization structure is particularly simple. It follows from Lemma~\ref{lem:logphi} that $\log\Phi$ is a concave function. Each summand in \eqref{eq:probit} is $\log\Phi$ composed with an affine function and is therefore concave~\cite[Chapter 3]{boyd}. Hence, $\widetilde\ell_H$ is concave on $\mathbb{R}\times(0,\infty)$. If the calibration sample contains at least two distinct reference margins, then $\widetilde\ell_H$ is strictly concave. Indeed, for two distinct parameter pairs $(\alpha_1,\beta_1)$ and $(\alpha_2,\beta_2)$, the affine values $\alpha_1+\beta_1\theta_i$ and $\alpha_2+\beta_2\theta_i$ cannot agree at two distinct values of $\theta_i$ unless the two parameter pairs coincide. Thus, at least one summand satisfies a strict concavity inequality, while all satisfy a nonstrict one. Proposition~\ref{prop:unique} then implies that any maximizer of \eqref{eq:probit} must be unique.\footnote{One can show that under reasonable assumptions, existence of a maximizer is guaranteed. Let $I_+=\{i:h_i=+1\}$ and $I_-=\{i:h_i=-1\}$, and suppose that both sets are nonempty. Define $\bar\theta_+=|I_+|^{-1}\sum_{i\in I_+}\theta_i$ and $\bar\theta_-=~|I_-|^{-1}\sum_{i\in I_-}\theta_i$. A sufficient condition for the existence of a maximizer of \eqref{eq:probit} is $\min_{i\in I_+}\theta_i<\max_{i\in I_-}\theta_i$ and $\bar\theta_+>\bar\theta_-$. This means that there should be some overlap between the two types of calls, but offside calls should nevertheless occur at larger reference margins on average than onside calls.} 

The requirement of distinct reference margins is important for identifying both parameters. If every reference margin equals $\theta_0$, then the likelihood depends on $(\alpha,\beta)$ only through $\alpha+\beta\theta_0$, or equivalently on $(\theta_0+m_H)/\sigma_H$, and the two human error parameters cannot be identified separately.

Problem \eqref{eq:probit} is a two-dimensional convex optimization problem and can be solved numerically using standard algorithms for convex optimization~\cite[Chapters~9--11]{boyd}. Since the feasible set $\mathbb{R}\times(0,\infty)$ is open, any maximizer $(\widehat\alpha,\widehat\beta)$ must satisfy the first-order conditions
\begin{align}
 \frac{\partial\widetilde\ell_{\HH}}{\partial\alpha}
 &=\sum_{i=1}^{n}h_i\lambda\bigl(h_i(\widehat\alpha+\widehat\beta\theta_i)\bigr)=0,
 \label{eq:probit-alpha}\\
 \frac{\partial\widetilde\ell_{\HH}}{\partial\beta}
 &=\sum_{i=1}^{n}h_i\theta_i\lambda\bigl(h_i(\widehat\alpha+\widehat\beta\theta_i)\bigr)=0,
 \label{eq:probit-beta}
\end{align}
where $\lambda(\cdot)$ is as in \eqref{eq:lambda}. Therefore, one can also solve these two nonlinear equations in the two unknowns $\widehat\alpha$ and $\widehat\beta$ directly. Under the distinct-margin condition above, any solution $(\widehat\alpha,\widehat\beta)$ with $\widehat\beta>0$ is the unique maximizer of \eqref{eq:probit}. We then recover the human error parameters as
\begin{equation}\label{eq:recover-human}
 \widehat m_{\HH}=\frac{\widehat\alpha}{\widehat\beta},\qquad
 \widehat\sigma_{\HH}=\frac1{\widehat\beta}.
\end{equation}

\Needspace{13\baselineskip}
\section{Summary and conclusions}\label{sec:conclusion}
We have developed a maximum-likelihood model for combining a human binary call
with an AI numerical measurement under independent Gaussian measurement errors.
We showed that the log-likelihood has a unique maximizer and that its derivative
at zero determines the sign of this maximizer. The resulting decision therefore
requires a threshold comparison rather than numerical optimization for each
incident. When both error means are zero, the threshold is
$\sqrt{2/\pi}\,\sdA^2/\sdH$; nonzero means give call-dependent thresholds and
an explicit correction to the AI measurement.

We also derived maximum-likelihood calibration procedures for the four error
parameters. The AI parameters have closed-form estimates, while the human
parameters are obtained from a two-dimensional concave maximization problem. These procedures require numerical reference margins, but
they do not require access to the human's internal measurements. Although we used offside calls to present the analysis throughout the paper, the same derivation applies whenever the human reports a binary decision about an underlying scalar quantity and the AI provides a numerical measurement of that same quantity. A promising direction for future work is to quantify the empirical impact of the proposed maximum-likelihood procedure on decision accuracy in applications where the model can be readily calibrated, such as finance, medical imaging, and prediction markets.

\section*{Acknowledgments}
Parsa and Radeen Alavi are grateful to their grandmother, Maryam Dara, for her support and encouragement throughout this project. The authors acknowledge the use of ChatGPT for generating figures, additional proofreading, and sanity checks, while following their $\sqrt{2/\pi}$ rule to ensure that AI did not overturn their human judgment. They take full responsibility for the content.



\end{document}